%% file: main.tex
\documentclass[runningheads]{llncs}

\usepackage{eccv}

\usepackage{eccvabbrv}

\usepackage{graphicx}
\usepackage{booktabs}

\usepackage[accsupp]{axessibility}  

\usepackage[pagebackref,breaklinks,colorlinks,citecolor=eccvblue]{hyperref}

\usepackage{orcidlink}

\usepackage{graphicx}
\usepackage{amsmath}
\usepackage{amssymb}
\usepackage{booktabs}
\usepackage{makecell}
\usepackage{multirow} 
\usepackage{multicol} 
\usepackage{algorithm}
\usepackage{algorithmic}
\usepackage{colortbl}  
\usepackage{xcolor}
\usepackage{array}   
\definecolor{Gray}{gray}{0.94}

\newcommand\highlight[1]{{\color{red}#1}}

\begin{document}

\title{{GlanceVAD}: Exploring \underline{Glance} Supervision for \\
Label-efficient \underline{V}ideo \underline{A}nomaly \underline{D}etection}
\newcommand{\modelname}{GlanceVAD~}

\titlerunning{GlanceVAD}

\author{Huaxin Zhang\inst{1} \and
        Xiang Wang\inst{1} \and
        Xiaohao Xu\inst{2} \and
        Xiaonan Huang\inst{2} \and
        Chuchu Han\inst{1} \and
        Yuehuan Wang\inst{1} \and
        Changxin Gao\inst{1} \and
        Shanjun Zhang\inst{3} \and
        Nong Sang\thanks{Corresponding author.}\inst{1} \\
        }

\authorrunning{Zhang et al.}


\institute{
Huazhong University of Science and Technology \and
University of Michigan, Ann Arbor \and 
Kanagawa Univerisity
}
\vspace{-4mm}

\maketitle

\input{src/0-abstract}

\input{src/1-introduction}

\input{src/2-related_work}

\input{src/3-proposed_method}

\input{src/4-experiments}

\input{src/5-conclusion}

\clearpage
\input{src/6-acknowledgement}

%
%
\bibliographystyle{splncs04}
\bibliography{main}

\clearpage
\appendix
\noindent\textbf{\Large Appendix}
\input{supply/appendix}

\end{document}

%% file: src/0-abstract.tex
\begin{abstract}
In recent years, video anomaly detection has been extensively investigated in both unsupervised and weakly supervised settings to alleviate costly temporal labeling.
Despite significant progress, these methods still suffer from unsatisfactory results such as numerous false alarms, primarily due to the absence of precise temporal anomaly annotation.
In this paper, we present a novel labeling paradigm, termed "glance annotation", to achieve a better balance between anomaly detection accuracy and annotation cost.
Specifically, glance annotation is a random frame within each abnormal event, which can be easily accessed and is cost-effective.
To assess its effectiveness, we manually annotate the glance annotations for two standard video anomaly detection datasets: UCF-Crime and XD-Violence.
Additionally,  we propose a customized \textbf{GlanceVAD} method, that leverages gaussian kernels as the basic unit to compose the temporal anomaly distribution, enabling the learning of diverse and robust anomaly representations from the glance annotations.
Through comprehensive analysis and experiments, we verify that the proposed labeling paradigm can achieve an excellent trade-off between annotation cost and model performance.
Extensive experimental results also demonstrate the effectiveness of our \textbf{GlanceVAD} approach, which significantly outperforms existing advanced unsupervised and weakly supervised methods.
Code and annotations will be publicly available at \href{https://github.com/pipixin321/GlanceVAD}{https://github.com/pipixin321/GlanceVAD}. 

\keywords{Video Anomaly Detection \and Weak Supervision \and Efficient Labeling}
\end{abstract}

%% file: src/1-introduction.tex
\section{Introduction}

\input{figs/teaser}
Video anomaly detection~\cite{ConvAE} aims to detect abnormal events in videos, holding significant potential for real-world tasks such as intelligent surveillance~\cite{ucf, xdviolence} and video content comprehension~\cite{huang2018makes,xu2022reliable,wang2023molo}. 
The fully-supervised setting~\cite{liu2019exploring,landi2019anomaly} is less explored because precise temporal boundary labeling of abnormal events is required, where annotators need to maintain focus and consistently adjust the video progress to patiently mark the boundaries, resulting in non-negligible labor costs and limiting its scalability.

To alleviate the prohibitive temporal anomaly labeling,
two typical paradigms have been widely adopted: unsupervised video anomaly detection~\cite{OCSVM,ConvAE,Luetal,GODs,stcgraph,generic_ad} and  weakly supervised video anomaly detection (WSVAD)~\cite{ucf,GCN,mist,xdviolence,rtfm,MSL,S3R,URDMU,UMIL}.
The former utilizes only normal videos for training to learn the common characteristics of normalcy. However, it is impossible for the normal videos in the training set to encompass all possible normal patterns, leading to the generation of numerous false alarms~\cite{cao2024survey}.
The latter leverages only video-level labels of normal and abnormal videos to explore anomalies and usually predicts the snippet-level anomaly score based on Multiple Instance Learning (MIL)~\cite{ucf} for its effectiveness.
Although recent weakly supervised works have achieved remarkable performance in many popular benchmarks, they still struggle to select reliable snippets for training due to the absence of location information for anomalous events, leading to significant bias towards the context of anomalies in the trained models. 
Therefore, a satisfactory balance between anomaly detection accuracy and annotation cost has not yet been achieved.

In this paper, inspired by recent studies~\cite{sf,lacp,hrpro,cui2022video,luo2023frequency,li2023d3g} that utilize label-efficient supervision, we explore a novel labeling paradigm in video anomaly detection called glance annotation.
As shown in \cref{fig:teaser}(a), annotators are required to randomly mark a frame within the time interval of the abnormal event.
Glance annotation offers several advantages, primarily demonstrated in the following aspects.
On the one hand, glance annotation provides a quick click in an anomaly instance, while frame-level annotation relies heavily on the expertise and judgment of the annotators, which can introduce subjectivity and potential inconsistencies in the annotations.
On the other hand, the annotation cost of glance annotation is almost comparable to weakly supervision, as both require watching the entire video to determine if there is an anomaly event, while the former can offer additional guidance information.

To explore the application of glance annotation in {video anomaly detection task,}
we design a standardized annotation process and manually annotate the two largest datasets for video anomaly detection, \ie, UCF-Crime~\cite{ucf} and XD-violence~\cite{xdviolence}.
Through statistics, we notice that glance annotations are very sparse, which is insufficient for the network to acquire robust representations of anomalous features.
Existing methods~\cite{mist,MSL} have explored generating binary pseudo-labels under weakly supervised settings. However, using 0/1 as supervision is challenging for modeling complex and diverse anomaly events. Inspired by the success of Gaussian Splatting in 3D scene representation~\cite{kerbl20233d,keselman2023flexible,wang2022voge}, we propose a novel Temporal Gaussian Splatting method to obtain smoother pseudo-labels.
It treats the video's anomaly score as the result of rendering different Gaussian anomaly kernels jointly.
During the training process, we initialize Gaussian anomaly kernels with glance annotations and continuously optimize the Gaussian anomaly kernels from the surroundings.
This stable and reliable pseudo-label effectively addresses the error accumulation caused by noise. 
Our proposed glance annotation and Temporal Gaussian Splatting can be inserted into existing MIL-based weakly-supervised methods.
Extensive experimental results demonstrate that our \textbf{GlanceVAD} can significantly outperform existing state-of-the-art video anomaly detection techniques and achieve an excellent trade-off between annotation cost and model performance (\cref{fig:teaser}(b)).

To summarize, our major contributions are as follows: 
\begin{itemize}
\item 
We introduce a new labeling paradigm for VAD task, termed glance annotation, which more closely mirrors real-world scenarios. We provide manual glance annotations for two largest video anomaly detection datasets, namely, UCF-Crime and XD-Violence.

\item 
We propose a novel Temporal Gaussian Splatting method, that performs reliable initialization and progressive updates of Gaussian anomaly kernels, to fully explore smooth and dense guidance from glance annotations. 

\item Our proposed \textbf{GlanceVAD}, based on several typical MIL-based networks, significantly outperforms baselines and  establishes state-of-the-art performance (\textbf{91.96\%}-AUC on UCF-Crime and \textbf{89.40\%}-AP on XD-Violence).
We also demonstrate the excellent trade-off between annotation cost and model performance, which can be flexibly adjusted in practical applications.
\end{itemize}

%% file: figs/teaser.tex
\begin{figure}[t]
\centering
\includegraphics[scale=0.43]{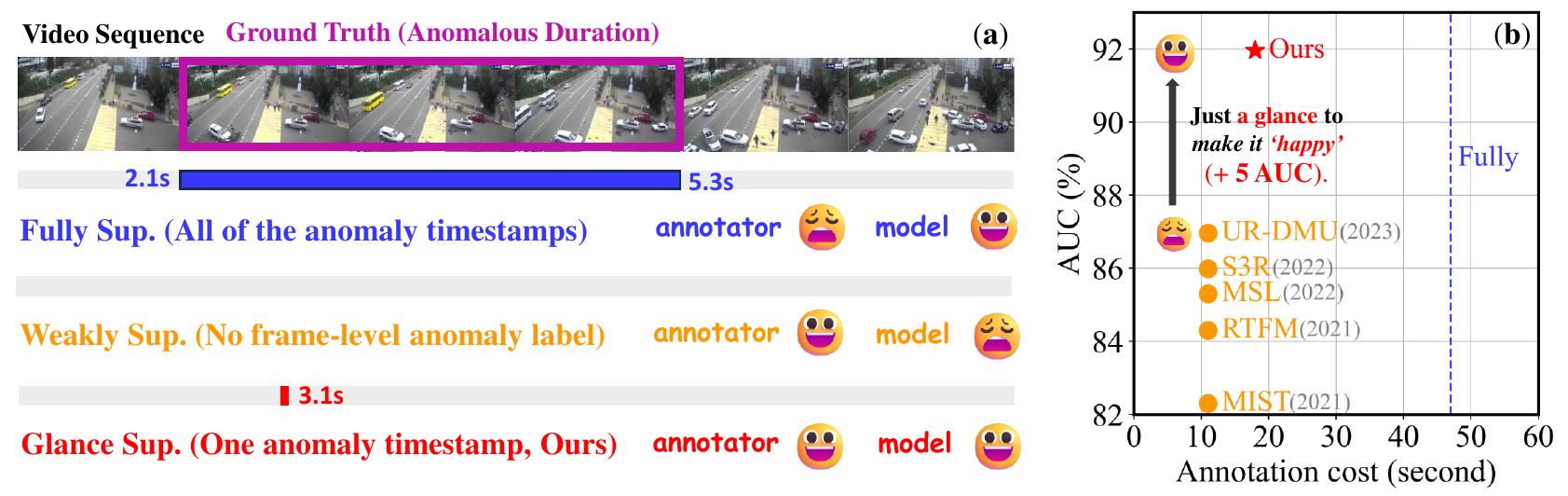}
\vspace{-6mm}
\caption{
\textbf{Motivation illustration}. {\bf{(a):}} Given an unlabelled video containing abnormal events, \textcolor{blue}{fully supervision} annotates all of the anomaly timestamps, \textcolor{orange}{weakly supervision} provides video-level labels (abnormal/normal), and the proposed \textbf{\textcolor{red}{glance supervision}} only requires annotating a single frame for each abnormal event.
{\bf{(b):}}
{We show that our proposed glance supervision exhibits excellent label efficiency, achieving leading performance over existing state-of-the-art weakly supervised methods (\eg, UR-DMU~\cite{URDMU}, S3R~\cite{S3R}, MSL~\cite{MSL}, \etc) while keeping labeling costs acceptable.}
}
\label{fig:teaser}
\vspace{-4mm}
\end{figure}

%% file: src/2-related_work.tex
\section{Related Works}

\noindent\textbf{Unsupervised video anomaly detection.}
Unsupervised methods treat anomaly detection as a task where only unlabeled normal samples are accessible during the training stage.
Traditional unsupervised methods~\cite{adam2008robust,kim2009observe,zhao2011online,mehran2009abnormal,Luetal,li2013anomaly} mainly utilize hand-crafted features and classical machine learning techniques.
Recently, methods based on deep models have become the mainstream of research.
These methods~\cite{ConvAE,rnn,framepred,gong2019memorizing,GODs,yang2023video} assume that normal frames can be well reconstructed or predicted, whereas abnormal frames yield substantial errors.
Specifically, Hasan~\etal~\cite{ConvAE} designed a fully connected neural network-based autoencoder to capture the temporal regularity in the video.
Liu~\etal~\cite{framepred} exploited the unpredictability of abnormal frames and designed a U-Net architecture~\cite{unet} for predicting future frames.
Despite the considerable advancements achieved by these methods, the normal videos in the training set fail to capture all normal patterns, leading to a significant number of false alarms during unforeseen normal events.

\vspace{1mm}
\noindent\textbf{Weakly supervised video anomaly detection.} 
Weakly supervised methods, which benefit from low annotation cost and incorporate both normal and abnormal training samples, have gained increasing attention in recent years.
Sultani~\etal~\cite{ucf} introduced a network based on Multiple Instance Learning (MIL).
Based on this, various customized solutions have been proposed.
Tian~\etal\cite{rtfm} explored the feature magnitude to facilitate anomaly detection and designed a Multi-scale Temporal Network (MTN) to capture multi-scale temporal dependencies.
Li~\etal~\cite{MSL} proposed to select consecutive snippets prediction with high anomaly scores.
Wu~\etal~\cite{S3R} and Zhou~\etal~\cite{URDMU} constructed memory banks to store prototypes of normal or abnormal snippets and enhanced the feature difference by interacting with memory.
Lv~\etal~\cite{UMIL} focused on eliminating variant context bias by seeking the invariant features across both confident and ambiguous groups.
Weakly supervised methods continue to experience context errors due to the absence of location information for abnormal frames.

\vspace{1mm}
\noindent\textbf{Point supervision.}
Point supervision is an efficient annotation paradigm that strikes a balance between labeling cost and model performance.
It was initially applied in image semantic segmentation~\cite{bearman2016s}, where each object  was annotated with a pixel instead of a mask.
Their pioneering research and success have inspired numerous studies in the field of video understanding,
including video recognition~\cite{mettes2016spot}, action localization~\cite{sf,lacp,hrpro}, temporal action segmentation~\cite{li2021temporal}, and video moment retrieval~\cite{cui2022video,li2023d3g}.
In these tasks, a single-frame click in the temporal dimension is provided for each action instance, offering an alternative to costly boundary annotations. 
To the best of our knowledge, we are the first to explore the application of point supervision, referred to as "glance annotation", in the domain of video anomaly detection.

\vspace{1mm}
\noindent\textbf{Gaussian Splatting.}
Recently, Gaussian Splatting has become a promising solution for 3D scene representation~\cite{kerbl20233d,keselman2023flexible,wang2022voge}.
They utilized 3D Gaussians as basic representation elements and render them into high-quality images with complex scenes through efficient splatting (a process from 3D Gussians to 2D image).
Inspired by their success, we apply this concept to video anomaly detection and use Gaussians as basic anomaly representation.
Different from the 3D Gaussian Splatting, we propose to initialize and update the Gaussian anomaly kernels on the temporal dimension.
Then we leverage the 1D anomaly score rendered by all Gaussian Kernels as the dense and smooth pseudo-labels for the model.

%% file: src/3-proposed_method.tex
\section{{Glance Supervision for Video Anomaly Detection }}
In this section, we start by introducing the task definition of video anomaly detection with glance supervision. Subsequently, we elucidate the process of constructing glance supervision for existing datasets.
Lastly, we present the baseline architecture employed in our study.

\noindent{{\textbf{Task definition.}}}
Video Anomaly Detection (VAD) aims to identify abnormal frames within an untrimmed video sequence $\mathcal{V}$. Existing weakly supervised methods commonly partition the video into multi-frame snippets. They employ a pretrained backbone, specifically I3D~\cite{i3d} to extract snippet-level features denoted as $X\in{\mathbb R}^{T\times D}$, followed by a detector $f_{\theta}(\cdot)$ to estimate abnormal scores for each snippet.
The detector is trained using both normal and abnormal videos with video-level labels $y$.
Here, $y=1$ indicates the presence of abnormal frames in the video, while $y=0$ indicates the absence of abnormalities.
In the context of weakly supervised settings, we explore a more informative and efficient labeling form referred to as glance supervision.
For each abnormal video in the training set, a randomly chosen single-frame click $g_i$ is provided between the start $s_i$ and end $e_i$ of each abnormal instance ($s_i<g_i<e_i$).
Notably, the glance annotation is marked on the temporal dimension.

\noindent{{\textbf{Proposed benchmarking dataset.}}}
To evaluate the effectiveness of glance supervision in video anomaly detection, we manually annotate glance annotations for the two largest existing datasets: UCF-crime~\cite{ucf} and XD-Violence~\cite{xdviolence}.
Unlike other video understanding tasks such as Video Moment Retrieval~\cite{cui2022video,li2023d3g} or Temporal Action Localization~\cite{lacp,hrpro}, anomalies in existing VAD datasets lack clear temporal boundaries and have a diverse temporal span.
Some anomalies, like car accidents, are very short-lived, while others, like fights and robberies, can be quite lengthy. This presents a great challenge for annotating anomaly videos, emphasizing the importance of designing a standardized glance annotation process.
\input{figs/glance_supervision}
In the annotation process, we strive to ensure the \textbf{Reliability} and \textbf{Comprehensiveness} of the glance annotation.
\textbf{Reliability} requires the annotated frame to be an anomaly, while \textbf{Comprehensiveness} demands that annotated frames cover as many abnormal events in the video as possible.
Based on these two principles, we have designed an annotation workflow:
annotators are required to quickly view the entire video, when there exists an anomaly event, they need to label a frame before the anomaly ends, and wait for the anomaly to end before annotating the next glance\footnote{More details about the annotation process can be found in Appendix.}.
The quantity and temporal distribution of the glance annotation are shown in the \cref{fig:glance_supervision}.

\input{figs/method_overview}
\noindent{{\textbf{Baseline architecture.}}}
The glance label is a form of weakly supervised label. 
Therefore, we take the mainstream approaches in weakly supervised setting as our baselines.
The majority of these methods are grounded in Multi Instance Learning (MIL), where the network predicts anomaly scores for each snippet, denoted as $A = f_{\theta}(X)$, then the top-k scores are averaged to obtain the video-level anomaly prediction $\hat{y}$.
\begin{equation}
\hat{y} = \frac{1}{K} \sum_{k\in topK(A)}^{K} A[k]
\end{equation}
Consequently, we can calculate cross-entropy loss by utilizing the video-level labels.
\begin{equation}
{\mathcal L}_{mil} = - ( ylog(\hat{y}) + (1-y)log(1-\hat{y}) )
\label{mil_loss}
\end{equation}
Recently, there have been numerous custom MIL-based methods proposed to improve the performance of VAD methods under weakly supervision. 
We select RTFM~\cite{rtfm} and UR-DMU~\cite{URDMU} as our baselines due to their straightforward implementation and outstanding performance, and we provide a comparison in \cref{sec:comparison_sota}.

{\section{Proposed Method: GlanceVAD}}
\noindent Glance annotation is inherently sparse in the temporal dimension and thus cannot serve as the sole basis for training a discriminative anomaly classifier.
Therefore, it is natural to consider mining additional reliable snippets around the glance annotations to provide denser guidance.
Existing weakly supervised approaches~\cite{mist,MSL} have explored generating snippet pseudo-labels based on snippet level anomaly scores and utilizing them to retrain the model. However, the improvement is limited due to the random initialization of the model and the error accumulation caused by unreliable pseudo labels.
The introduction of glance supervision can address the above issues. 
However, a question still remains about how to effectively initialize and update pseudo labels in the presence of glance annotation.
As shown in \cref{fig:method_overview}, inspired by the 3D Gaussian representation in~\cite{kerbl20233d}, we adopt the concept of Gaussian Splatting and incorporate Temporal Gaussian Splatting in the video anomaly detection domain. Temporal Gaussian Splatting enables reliable initialization and progressive updates of Gaussian anomaly kernels during the training process. Then we leverage the 1D anomaly score rendered by
all kernels as the dense and smooth pseudo-labels for the model.

\vspace{4mm}
\subsection{{Temporal Gaussian Anomaly Representation}}
Binary pseudo labels~\cite{mist,MSL} simply set the label values of snippets to 0 and 1. 
However, real-world anomalies often exhibit gradual changes and exert varying degrees of influence on the video context.
Taking into account these characteristics of anomalies, we use three values to parameterize each Gaussian anomaly kernel: $\mathbf{\mu}_i$ represents the current temporal position, $o_i \in [0,1]$ indicates the severity of the anomaly, and $r_i$ represents the context radius.
Each Gaussian anomaly kernel influences a temporal point $\mathbf{t}\in [1, T]$ according to the standard Gaussian equation weighted by its severity:
\begin{equation}
f_i(\mathbf{t}) = o_i \cdot exp(-\frac{\|\mathbf{t}-\mathbf{\mu_i}\|^2}{2r_i^2})
\label{eq:gaussian_represantation}
\end{equation}
Assuming that the video has an infinite sequence of Gaussian anomaly kernels from start to end, splatting all the Gaussian anomaly kernels can render anomaly scores at different time points in the video.
\begin{equation}
\hat{A}(\mathbf{t}) = \textbf{norm}(\sum_{i=1}^T{f_i(\mathbf{t})})
\label{eq:splatting}
\end{equation}
In our implementation, we fixed the number of Gaussian anomaly kernels to be equal to the number of video snippets.

\subsection{{Gaussian Kernel Initialization}}
We leverage the extremely weak and cost-efficient glance-level labels for the initialization of the Gaussian anomaly kernels.
Since glance annotation can capture the attention of annotators, we assume that glance annotation has a high degree of anomaly severity ($o_i$) and a significant impact on the context.
\begin{equation}
o_{i}=\left\{\begin{array}{ll}
1 & \text { if } \quad \mu_{i} \in \mathcal{G} \\
0 & \text { others }
\end{array}\right.
\end{equation}
$\mathcal{G}$ is the set that contains all glance annotations in the input video.
The context radius $r_i$ of each Gaussian anomaly kernel in \cref{eq:gaussian_represantation} with glance label is set to a fixed value $r_g$.

\input{tables/alg_alm}
\subsection{{Dynamic Gaussian Kernel Optimization}}
\textbf{Gaussian Kernel Mining}
The anomaly Gaussian kernels are updated during each training iteration so that the rendered anomaly score can more closely align with the ground truth anomaly distribution in the video.
To achieve this, we propose Gaussian Kernel Mining to mine pseudo anomaly snippets $\mathcal{T}^a$ with high confidence around glance annotation. 
As illustrated in \cref{alg:alm}, we employ a dynamic threshold and perform local bidirectional mining based on the glance annotations to avoid insufficient pseudo samples during the early stages of training when anomaly scores are initially low.
Snippets with anomaly scores exceeding a specific proportion of the glance snippet score are identified as pseudo anomaly snippets.
Their corresponding Gaussian anomaly kernel is updated as follow:
\begin{equation}
o_{i}=\left\{\begin{array}{ll}
1 & \text { if } \quad \mu_{i} \in \mathcal{G} \cup \mathcal{T}^a \\
0 & \text { others }
\end{array}\right.
\end{equation}

\noindent\textbf{Differentiable rendering via splatting.} 
Given the weak supervision provided by glance annotation, it is challenging and redundant to force the network to predict the Gaussian kernel for each snippet.
Therefore, we render Gaussian anomaly kernels into a 1D vector through \cref{eq:splatting}, where each element indicates the anomaly score for each snippet. 
The rendered anomaly score is then utilized to supervise the anomaly score output by the model.
\begin{equation}
{\mathcal L}_{abn} = BCE(A(t), \hat{A}(\mathbf{t}))
\end{equation}

\subsection{Training and Inference Details} 
\textbf{Training.}
In order to balance the number of normal and abnormal samples, we follow previous works~\cite{rtfm,S3R,URDMU} to construct normal/abnormal video pairs and input them into the model.
The traditional MIL loss ${\mathcal L}_{nor}$ is adopted to normal videos.
Finally, the training loss is constructed as follows:
\begin{equation}
{\mathcal L}_{total} = {\mathcal L}_{base} + {\mathcal L}_{abn} + {\mathcal L}_{nor}
\end{equation}
where ${\mathcal L}_{base}$ indicates the loss function in our baseline methods\footnote{More details about our baselines can be found in Appendix.}.
We adopt different baselines and compare them in the next experimental section.

\noindent\textbf{Inference.}
Following existing methods~\cite{URDMU,rtfm,S3R},
we input the testing video into the model to obtain the snippet-level anomaly score, and then convert it into the final frame-level anomaly score through repetition.

%% file: figs/glance_supervision.tex
\begin{figure}[t]
\centering
\includegraphics[scale=0.35]{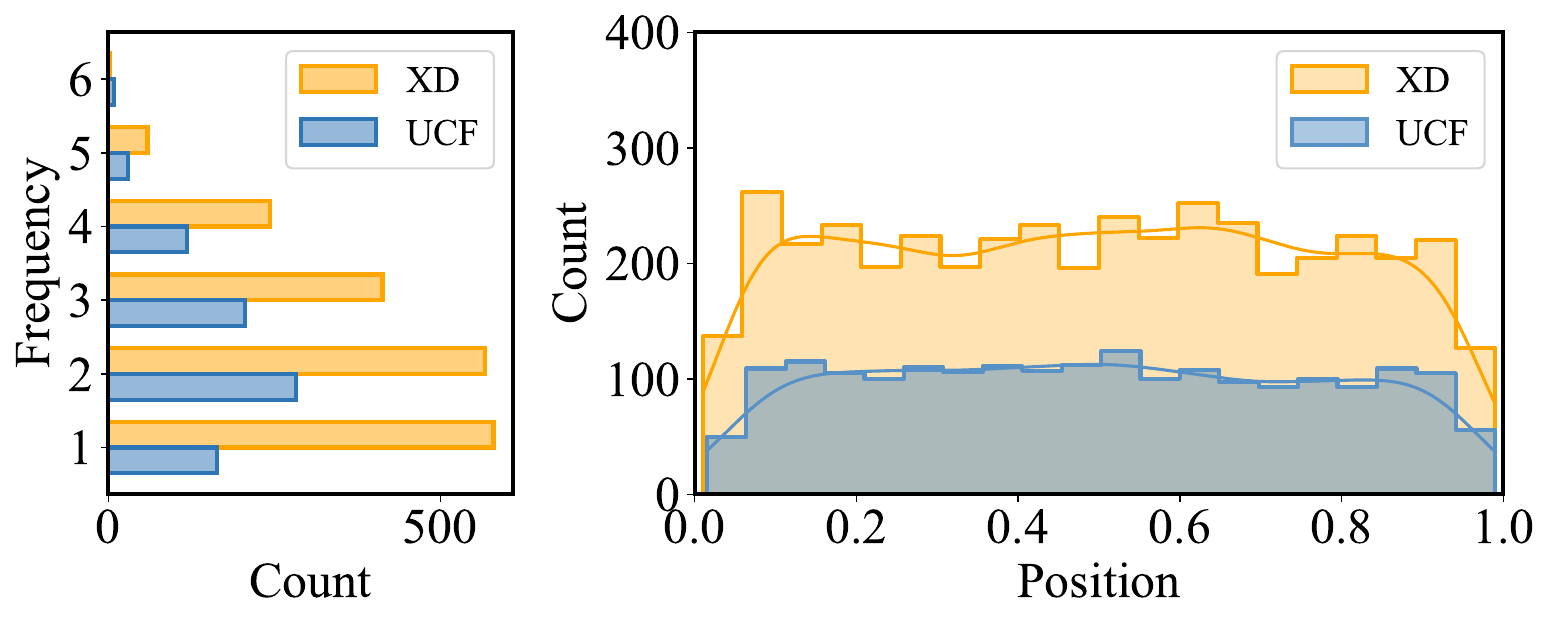} 
\vspace{-2mm}
\caption{
\textbf{Overall statistical results of glance annotations on the XD-Violence~\cite{xdviolence} and UCF-Crime~\cite{ucf} datasets}. The left figure illustrates the frequency statistics of glance annotations within the video, while the right figure depicts the temporal position distribution of glance annotations.
}
\vspace{-4mm}
\label{fig:glance_supervision}
\end{figure}

%% file: figs/method_overview.tex
\begin{figure*}[t]
\centering
\includegraphics[width=\textwidth]{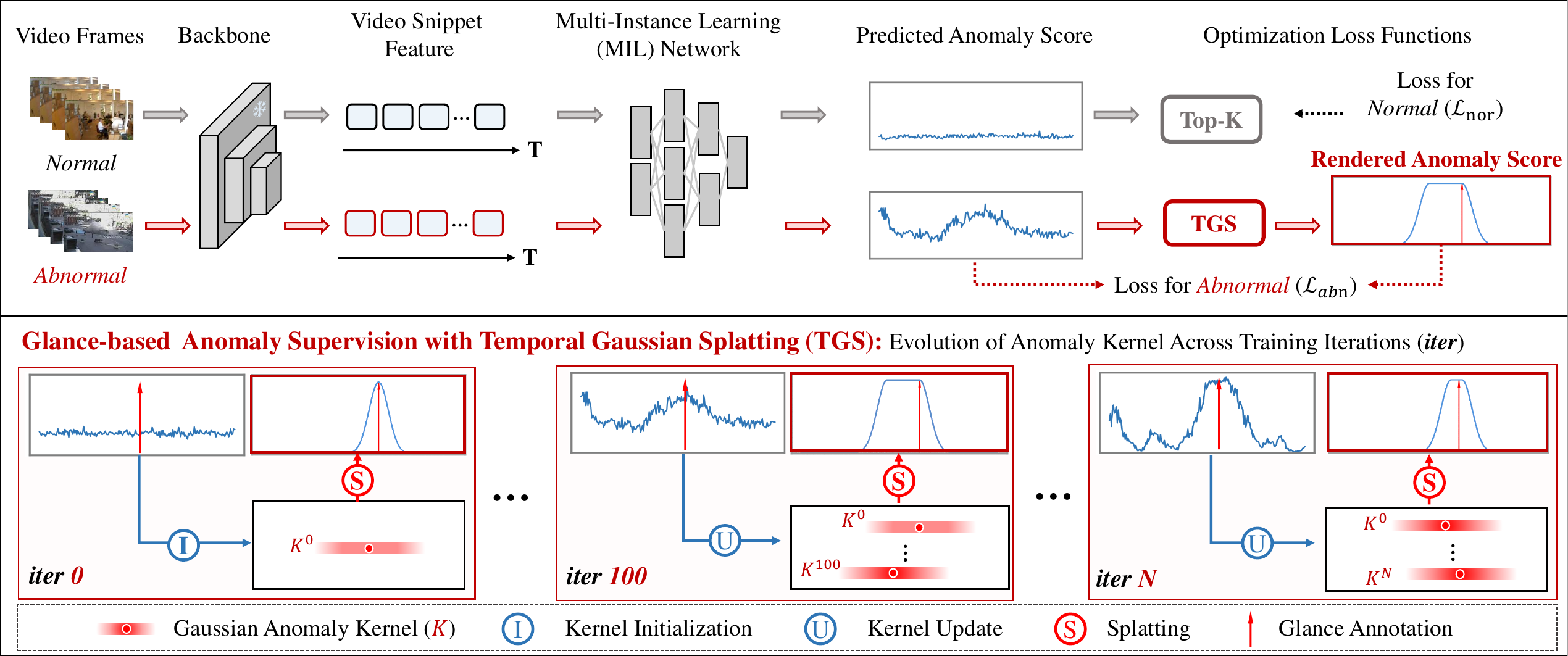}
\vspace{-4mm}
\caption{\textbf{Overview of the proposed method.} We use a pretrained video feature extraction network,\ie, I3D~\cite{i3d}, to extract snippet-level features from the videos. 
These features are then input to existing MIL-based methods (\eg, MIL~\cite{ucf}, RFTM~\cite{rtfm}, UR-DMU~\cite{URDMU}) to obtain anomaly scores.
{We propose Temporal Gaussian Splatting to generate pseudo-labels to supervise the anomaly score of abnormal videos.}
}
\label{fig:method_overview}
\end{figure*}

%% file: tables/alg_alm.tex
\begin{algorithm}[h]
\caption{Gaussian Kernel Mining.}
\label{alg:alm}
\raggedright
\textbf{Input}: Anomaly score $\mathbf{A}\in{\mathbb R}^T$, glance annotations $\mathcal{G}=\{g_i\}_i^{N_g}$, anomaly ratio $\alpha$.\\
\textbf{Output}: Pseudo anomaly snippets $\mathcal{T}^a=\{t_i\}_i^{N_{a}}$.

\begin{algorithmic}[1] 
\STATE Let $\mathcal{T}^a \gets \varnothing$.
\FOR{every $g_i \in \mathbf{G}$}
    \FOR{$t = g_i$ {\bfseries to} $g_{i-1}$}
        \STATE \textbf{if} $\mathbf{A}[t] > \alpha \cdot A[g_i] $, \textbf{then} $\mathcal{T}^a \gets t \cup \mathcal{T}^a$, \textbf{else break}
        \STATE \textbf{end if}
    \ENDFOR

    \FOR{$t = g_i$ {\bfseries to} $g_{i+1}$}
        \STATE \textbf{if} $\mathbf{A}[t] > \alpha \cdot A[g_i]$, \textbf{then} $\mathcal{T}^a \gets t \cup \mathcal{T}^a$, \textbf{else break}
        \STATE \textbf{end if}
    \ENDFOR
\ENDFOR
\STATE \textbf{Return} $ \mathcal{T}^a$
\end{algorithmic}
\end{algorithm}

%% file: src/4-experiments.tex
\section{Experiments}
\subsection{Experimental Setup}

\noindent\textbf{Datasets}. 
We conduct our experiments on two standard video anomaly detection datasets, namely, UCF-Crime~\cite{ucf} and XD-Violence~\cite{xdviolence}. 
Notably, following the data partitioning approach in previous weakly supervised setting, we manually provided glance annotations for each abnormal video in the training set.

\textbf{1}) \textbf{UCF-Crime} is a large-scale anomaly detection dataset that contains 1900  untrimmed videos with a total duration of 128 hours from both outdoor and indoor surveillance cameras. 
It covers 13 classes of real-world anomalies, such as \textit{Abuse}, \textit{Explosion}, \textit{Fighting}, and \textit{Shooting}. 
The number of videos for training/testing is 1610/290 in the weakly-supervised setting, and the training set contains 810 abnormal videos and 800 normal videos, respectively.
After re-labeling, we collected an average of 2.50 single-frame annotations per video.

\textbf{2}) \textbf{XD-Violence} is currently the largest benchmark which contains 4754 videos with a total duration of 217 hours from various sources, such as surveillance, movies, car cameras, and games. 
It contains 6 classes of anomalies including \textit{Abuse}, \textit{Car Accidents}, \textit{Explosions}, \textit{Fighting}, \textit{Riots}, and \textit{Shooting}. 
Unlike UCF-Crime, which features fixed-camera videos, XD-Violence consists of numerous videos recorded by moving cameras. 
The number of videos for training/testing is 3954/800, following the weakly-supervised setting. The training set contains 1905 abnormal videos and 2049 normal videos respectively.
After re-labeling, we collected an average of 2.22 single-frame annotations per video.

\noindent\textbf{Evaluation metric}.
Following previous works~\cite{ucf, rtfm, S3R, URDMU}, 
we leverage
the Area Under the Curve (AUC) of the frame-level Receiver Operating Characteristic (ROC) as the main
evaluation metric for UCF-Crime. 
Meanwhile, the AUC of the frame-level precision-recall curve (AP) is utilized for XD-Violence as the standard evaluation metric. 
A higher AUC/AP indicates a greater margin between the predictions of normal and abnormal snippets, suggesting a more effective anomaly classifier.
Inspired by~\cite{UMIL, vadclip}, we also evaluate the AUC/AP of abnormal videos (termed by AUC$_A$/AP$_A$) to uncover the poor accuracy on abnormal videos that concealed by superior performance on normal videos in the test set.

\noindent\textbf{Implementation details.}
For a fair comparison, we follow existing methods~\cite{rtfm,S3R,URDMU}, dividing each video into 16-frame snippets and utilizing I3D network, which is pretrained on Kinetics-400~\cite{i3d}, as the feature extractor.
We also follow them to adopt crop augmentation strategy, which includes using the center, four
corners, and their mirrored counterparts for UCF-Crime and the center, four
corners for XD-Violence.
In the training stage, we resample each video into N = 200 snippets for batch training.
We employ the Adam optimizer with a learning rate of 1e-4 and a weight decay of 5e-4,
and set the batch size as 64.
Hyper-parameters, including $\alpha=0.9$, $r_g =0.1$, were set by grid search, and their effects are experimentally analyzed in \cref{sec:ablation}.

\subsection{Comparison with State-of-The-Art Methods}
\label{sec:comparison_sota}
We evaluate the effectiveness of our proposed method by comparing it against the most recent unsupervised and weakly supervised video anomaly detection methods.
Our proposed method can be applied as a plug-and-play module to various existing weakly supervised methods. 
To demonstrate the generality of our approach, we conducted experiments on different baselines including MIL~\cite{ucf}, RTFM~\cite{rtfm}, and UR-DMU~\cite{URDMU}. 

\input{tables/comparision_ucf}
\input{tables/comparision_xd}
\noindent\textbf{Results of UCF-Crime and XD-Violence.}
Compared to previous state-of-the-art methods in \cref{tab:comparision_ucf}, our proposed method achieves state-of-the-art performance on UCF-Crime testing set and shows significant improvement across different baselines.
Our method has an AUC of 91.96\%, outperforming the prior state-of-the-art method UR-DMU~\cite{URDMU} by 4.99\%.
Notably, our method demonstrates superior performance in AUC of abnormal videos (84.94\% vs 70.81\%),
which demonstrates that our method can more accurately locate the abnormal parts in the abnormal videos.
As shown in \cref{tab:comparision_xd}, our proposed method also build a new state-of-the-art performance on XD-Violence testing set.
Our method has an AP of 89.40\%, outperforming the prior SoTA method~\cite{MACILSD} which also use audio input by 6.00\%.
The same significant improvement in abnormal videos is also observed in XD-Violence (89.85\% vs 83.94\%)

\subsection{Ablation Study}
\label{sec:ablation}
We conduct comprehensive ablation studies on multiple benchmarks to investigate the capability of the proposed \modelname and analyze the role of each component. Unless otherwise specified, \modelname based on UR-DMU~\cite{URDMU} is adopted as the default setting for ablation.

\input{tables/ablation_component}
\noindent\textbf{Model component effectiveness study.}
To investigate the impact of various components, we conduct experiments to ablate each component detailed in \cref{tab:ablation_component}.
The \textit{Baseline} refers to the re-implementation of UR-DMU, achieving an AUC of 86.54\% on UCF-Crime and an AP of 82.34\% on XD-Violence.
With the inclusion of \textit{Mining}, the performance improves to 86.94\% and 86.43\%. 
After introducing dynamic anomaly score thresholds, the results further improves to 90.38\% and 87.65\%. 
This demonstrates that a dynamic threshold can help to mine low-confidence snippets around the glance annotations in the early training stage. 
By introducing Gaussian Splatting, the results reach 90.89\% and 88.89\%, we infer that this is because the anomaly score generated by Gaussian Splatting better preserves the discriminative ability for different degrees of anomalies compared to binary labels, enabling the model to learn more diverse representations of anomalies. 
Upon incorporating all components that complement each other, the performance further boost to 91.96\% and 89.40\%.

\input{figs/params}
\noindent\textbf{Influence of hyper-parameters $r_g$ and $\sigma$.}
We conduct ablation experiments on the parameters $r_g$ and $\sigma$ to investigate their impact on performance, as shown in \cref{fig:para}.
The best results are obtained when $r_g=0.1$ and $\alpha=0.9$ on both datasets.
The performance starts to drop as $r_g$ increases and $\beta$ descends due to the fact that more unreliable snippets are mined as pseudo anomaly snippets.
We find that the model performance on UCF-Crime is more sensitive to the parameters.
This phenomenon can be attributed to the unchanging nature of scenes in the UCF-Crime dataset,
which leads to higher similarity in snippet features within the video, resulting in an increased risk of excessive mining.

\input{tables/ablation_glance_position}
\noindent\textbf{Influence of the position of glance annotations.}
To assess the robustness of our method to the temporal position of glance annotation, we introduced varying degrees of random perturbations to the original positions of the annotated glances. 
The results are presented in the \cref{tab:ablation_glance_position}.
We find that our method shows no significant performance degradation after applying random perturbations to the glance positions. This indicates that our method possesses a notable tolerance towards variations in glance annotation positions.

\input{tables/ablation_training}
\noindent\textbf{Trade-off between annotation cost and performance.}
The number of abnormal and normal videos in the training set is nearly equal in existing datasets such as UCF-Crime and XD-Violence. However, collecting such a large number of abnormal videos is extremely challenging in practical applications. 
Therefore, we conduct comparative experiments with varying proportions of anomaly training videos with weakly supervision and glance supervision, as shown in the \cref{tab:ablation_training}.
The result suggests that our method provides an effective trade-off between data/annotation cost and model performance. 

\input{tables/ablation_gaussian_kernel}
\noindent\textbf{Influence of Gaussian kernels with
 different distributions.}
We conducted experiments to assess the impact of Gaussian anomaly kernels with different distributions, \ie, \textit{Normal}, \textit{Cauchy}, and \textit{Laplace}.
The results are presented in \cref{tab:gaussian_kernel}.
It is observed that the Gaussian kernel with a normal distribution outperforms the other two distributions.
We attribute this to the faster tail decay and smoother central neighborhood of the Gaussian kernel with a normal distribution. This characteristics provides an advantage in exploring more reliable anomaly kernels.

\input{figs/quality_result}
\subsection{Qualitative Results}
In \cref{fig:quality}, we compare our \modelname with UR-DMU~\cite{URDMU} for video anomaly detection on test videos in UCF-Crime. Our model shows more accurate detection of anomaly frames.
Specifically, our method effectively distinguishes between anomaly and normal frames, mitigating false anomaly predictions - a challenge that UR-DMU struggles with.
This reveals that glance supervision can enable the model to focus on the anomaly part of the video and reduce significant bias towards the confusing context frames.

%% file: tables/comparision_ucf.tex
\begin{table}[t]
\centering
\caption{Comparison of the frame-level performance AUC on UCF-Crime testing set with existing approaches. \highlight{$\uparrow$} denotes the relative performance gain between our method and the baseline method. “$\ast$” represents the result of our implementation.}
\label{tab:comparision_ucf}
\vspace{-2mm}
\resizebox{0.9\textwidth}{!}{
\begin{tabular}{>{\centering}p{3cm}|p{4cm}c|>{\centering}p{3cm}c}
\hline\hline
\textbf{Supervision} & \textbf{Methods}  & \textbf{Feature} & \textbf{AUC(\%)} & \textbf{AUC$_A$(\%)}  \\
\hline
\multirow{4}{*}{\begin{tabular}{c}Unsupervised\end{tabular}}
& Conv-AE~\cite{ConvAE}  & - & 50.60 & N/A \\
& Lu~\etal~\cite{Luetal}  & C3D & 65.51 & N/A\\
& GODS~\cite{GODs} & I3D & 70.46 & N/A\\
& STC Graph~\cite{stcgraph}  & RPN & 72.70 & N/A\\
\hline
\multirow{13}{*}{\begin{tabular}{c}Weakly \\ supervised\end{tabular}}
& Sultani~\etal~\cite{ucf} & C3D & 75.41 & N/A\\
& GCN-Anomaly~\cite{GCN} & TSN & 82.12 & N/A\\
& MIST~\cite{mist}  & I3D & 82.30 & N/A\\
& Wu~\etal~\cite{xdviolence}  & I3D & 82.44 & N/A\\
& RTFM~\cite{rtfm}  & I3D & 84.30 & 63.86\\
& MSL~\cite{MSL}  & I3D & 85.30 & N/A\\
& S3R~\cite{S3R}  & I3D & 85.99 & N/A\\
& UMIL~\cite{UMIL} & X-CLIP & 86.75 & 68.68\\
& UR-DMU~\cite{URDMU}  & I3D & \underline{86.97} & \underline{70.81}\\
& MIL$^*$~\cite{ucf} & I3D &80.83 & 62.27 \\
& RTFM$^*$~\cite{rtfm}  & I3D & 83.50 & 62.96\\
& UR-DMU$^*$~\cite{URDMU}  & I3D & 86.54 & 70.52\\
\hline
\multirow{3}{*}{\begin{tabular}{c}Glance \\ supervised\end{tabular}}
 &  \cellcolor{Gray}\textbf{GlanceVAD (MIL)} &  \cellcolor{Gray}I3D &  \cellcolor{Gray}87.30\highlight{$^{\uparrow6.47}$} & \cellcolor{Gray}71.07\highlight{$^{\uparrow8.80}$} \\
& \cellcolor{Gray}\textbf{GlanceVAD (RTFM)}  &  \cellcolor{Gray}I3D &  \cellcolor{Gray}87.80\highlight{$^{\uparrow4.30}$} & \cellcolor{Gray}75.16\highlight{$^{\uparrow12.20}$}\\
& \cellcolor{Gray}\textbf{GlanceVAD (UR-DMU)}  &  \cellcolor{Gray}I3D &  \cellcolor{Gray}\textbf{91.96}\highlight{$^{\uparrow5.42}$} & \cellcolor{Gray}\textbf{84.94}\highlight{$^{\uparrow14.42}$}\\
\hline\hline
\end{tabular}
}
\end{table}

%% file: tables/comparision_xd.tex
\begin{table}[t]
\centering
\caption{Comparison of frame-level performance AP on XD-Violence validation set. \highlight{$\uparrow$} denotes the relative performance gain between our method and the baseline method. “$\ast$” represents the result of our implementation.}
\label{tab:comparision_xd}
\vspace{-2mm}
\resizebox{0.9\textwidth}{!}{
\begin{tabular}{>{\centering}p{3cm}|p{4cm}c|>{\centering}p{3cm}c}
\hline\hline
\textbf{Supervision} & \textbf{Methods}  & \textbf{Feature} & \textbf{AP(\%)} & \textbf{AP$_A$(\%)}\\
\hline
\multirow{2}{*}{\begin{tabular}{c}Unsupervised\end{tabular}}
& OCSVM~\cite{OCSVM}  & - & 27.25 & N/A\\
& Conv-AE~\cite{ConvAE}  & - & 30.77 & N/A\\
\hline
\multirow{12}{*}{\begin{tabular}{c}Weakly \\ supervised\end{tabular}}
& Sultani~\etal~\cite{ucf}  & I3D & 75.68 & N/A\\
& Wu~\etal~\cite{xdviolence}  & I3D+VGGish & 78.64 & N/A\\
& RTFM~\cite{rtfm}  & I3D & 77.81 & N/A\\
& MSL~\cite{MSL}  & I3D & 78.28 & N/A\\
& S3R~\cite{S3R}  & I3D & 80.26 & N/A\\
& UR-DMU~\cite{URDMU}  & I3D & 81.66 & 83.51\\
& UR-DMU~\cite{URDMU}  & I3D+VGGish & 81.77 & N/A\\
& MACIL-SD~\cite{MACILSD}  & I3D+VGGish & \underline{83.40} & N/A\\
& MIL$^*$~\cite{ucf} & I3D & 75.24 & 78.61 \\
& RTFM$^*$~\cite{rtfm}  & I3D & 77.53 & 78.57\\
& UR-DMU$^*$~\cite{URDMU}  & I3D & 82.40 & \underline{83.94}\\
\hline
\multirow{3}{*}{\begin{tabular}{c}Glance \\ supervised\end{tabular}}
& \cellcolor{Gray}\textbf{GlanceVAD (MIL)}  &  \cellcolor{Gray}I3D & \cellcolor{Gray}83.61\highlight{$^{\uparrow8.37}$} & \cellcolor{Gray}86.15\highlight{$^{\uparrow7.54}$}\\
& \cellcolor{Gray}\textbf{GlanceVAD (RTFM)}  &  \cellcolor{Gray}I3D & \cellcolor{Gray}86.88\highlight{$^{\uparrow9.35}$} & \cellcolor{Gray}86.65\highlight{$^{\uparrow8.08}$}\\
& \cellcolor{Gray}\textbf{GlanceVAD (UR-DMU)}  &  \cellcolor{Gray}I3D & \cellcolor{Gray}\textbf{89.40}\highlight{$^{\uparrow7.00}$} & \cellcolor{Gray}\textbf{89.85}\highlight{$^{\uparrow5.91}$}\\
\hline\hline
\end{tabular}
}
\end{table}

%% file: tables/ablation_component.tex
\begin{table}[!t]
\centering
\caption{Ablation study of our proposed Temporal Gaussian Splatting on UCF-Crime and XD-Violence.
We use UR-DMU as our baseline in this table.
\textit{Mining}: Gaussian Kernel Mining is adopted to expand pseudo anomaly samples.
\textit{Dynamic}: Dynamic threshold is used in Gaussian Kernel Mining.
\textit{Gaussian}: Gaussian Splatting is conducted to each glance annotation and pseudo anomaly snippet.}
\label{tab:ablation_component}
\vspace{-2mm}
\resizebox{0.8\textwidth}{!}{
\begin{tabular}{>{\centering}p{1.5cm}>{\centering}p{1.5cm}>{\centering}p{1.5cm}>{\centering}p{1.5cm}|>{\centering}p{3cm}c}
\hline\hline
Baseline & Mining & Dynamic & Gaussian & \textbf{AUC(\%)-UCF} & \textbf{AP(\%)-XD} \\
\hline
$\checkmark$ &   &   &  & 86.54 & 82.34 \\
$\checkmark$ & $\checkmark$  &   &  & 86.94 & 86.43 \\
$\checkmark$ &  &   & $\checkmark$  & 88.06 & 86.84 \\
$\checkmark$ & $\checkmark$  & $\checkmark$  &  & 90.38 & 87.65 \\
$\checkmark$ & $\checkmark$  &   & $\checkmark$ & 90.89 & 88.89 \\
$\checkmark$ & $\checkmark$   & $\checkmark$   & $\checkmark$  & \textbf{91.96} & \textbf{89.40} \\
\hline\hline
\end{tabular}
}

\end{table}

%% file: figs/params.tex
\begin{figure}[t]
  \centering\begin{subfigure}[t]{0.4\linewidth}
    \centering
    \includegraphics[width=\linewidth]{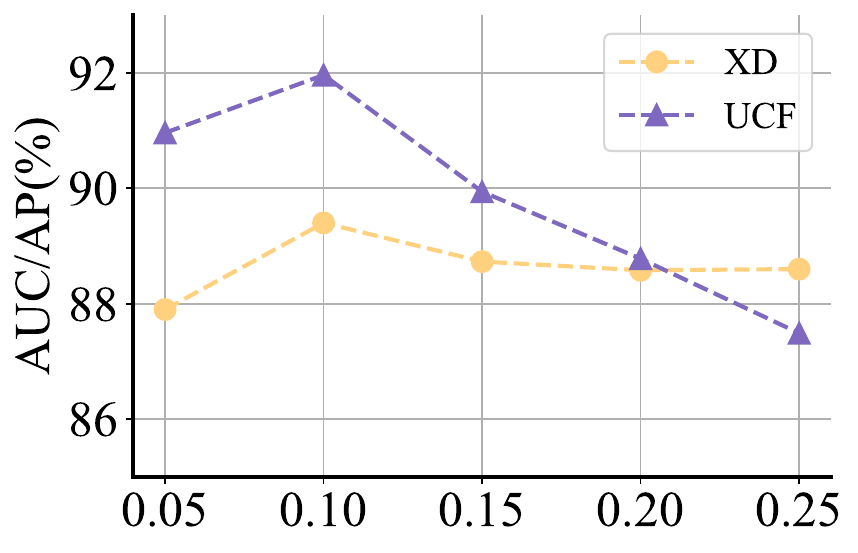}
    \caption{$r_g$}
  \end{subfigure}\begin{subfigure}[t]{0.4\linewidth}
    \centering
    \includegraphics[width=\linewidth]{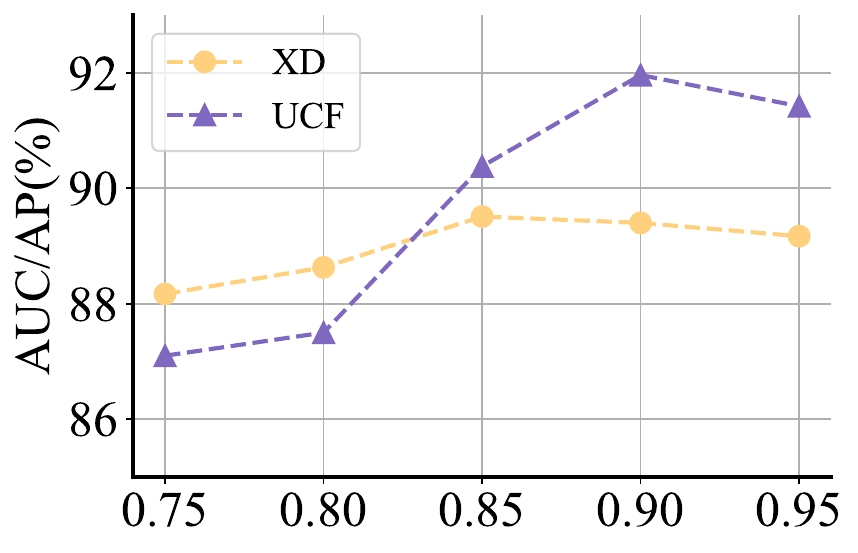}
    \caption{$\alpha$}
  \end{subfigure}
  \vspace{-3mm}
  \caption{Parameter analysis of $r_g$ and $\alpha$ on UCF-Crime and XD-Violence.}
  \label{fig:para}
\end{figure}

%% file: tables/ablation_glance_position.tex
\begin{table}[!t]
\centering
\caption{Results of different degree of random temporal perturbations to the annotated glances.}
\label{tab:ablation_glance_position}
\vspace{-2mm}
\resizebox{0.8\textwidth}{!}{
\begin{tabular}{>{\centering}p{3cm}|>{\centering}p{2cm}>{\centering}p{2cm}|>{\centering}p{2cm}c}
\hline\hline
\multirow{2}{*}{Perturbation(frame)} & \multicolumn{2}{c|}{UCF-Crime} & \multicolumn{2}{c}{XD-Violence} \\
& \textbf{AUC(\%)} & \textbf{AUC$_A$(\%)}  & \textbf{AP(\%)} & \textbf{AP$_A$(\%)}\\
\hline
0 & \textbf{91.96} & \textbf{84.94} & \textbf{89.40} & \textbf{89.85} \\
10 & 91.55 & 84.44 & 89.37 & 89.28 \\
50 & 90.84 & 82.38 & 89.07 & 89.12 \\
100 & 90.43 & 80.28 & 88.90 & 88.98 \\
\hline\hline
\end{tabular}
}

\end{table}

%% file: tables/ablation_training.tex
\begin{table}[!t]
\centering
\caption{Ablation study of the ratio of training abnormal videos with different supervision. W.(Weakly supervision), G.(Glance supervision)}
\label{tab:ablation_training}
\vspace{-2mm}
\resizebox{0.8\textwidth}{!}{
\begin{tabular}{>{\centering}p{3cm}cc|cc|cc}
\hline\hline
\multirow{2}{*}{Settings} & \multirow{2}{*}{W.} & \multirow{2}{*}{G.} & \multicolumn{2}{c|}{UCF-Crime} & \multicolumn{2}{c}{XD-Violence} \\
& & & \textbf{AUC(\%)} & \textbf{AUC$_A$(\%)} & \textbf{AP(\%)} & \textbf{AP$_A$(\%)} \\
\hline
\textit{Weakly (Full)} & \cellcolor{orange!100}100\% & \cellcolor{gray!10}0\% & 86.54 & 70.52 &82.40 & 83.94 \\
\hline
\multirow{3}{*}{\begin{tabular}{c}\textit{Weakly (Part)}\end{tabular}}
& \cellcolor{orange!25}25\% & \cellcolor{gray!10}0\%  &  82.56 & 63.25 & 78.46 & 78.73 \\
& \cellcolor{orange!50}50\% & \cellcolor{gray!10}0\%  & 83.15 & 64.96 & 79.22 & 79.36 \\
& \cellcolor{orange!75}75\% & \cellcolor{gray!10}0\%  & 85.64 & 67.54 & 80.22 & 80.16 \\

\hline
\multirow{3}{*}{\begin{tabular}{c}\textit{Glance (Part)}\end{tabular}}
& \cellcolor{gray!10}0\% & \cellcolor{red!15}25\%  & 88.43 & 76.24 & 87.75 & 88.20 \\
& \cellcolor{gray!10}0\% & \cellcolor{red!30}50\%  & 90.03 & 78.54 & 88.57 & 88.56 \\
& \cellcolor{gray!10}0\% & \cellcolor{red!45}75\%  & 90.56 & 80.43 & 88.95 & 89.76 \\

\hline
\multirow{3}{*}{\begin{tabular}{c}\textit{Mixed}\end{tabular}}
& \cellcolor{orange!75}75\% & \cellcolor{red!15}25\%  & 89.06 & 76.52 & 88.23 & 88.35 \\
& \cellcolor{orange!50}50\% & \cellcolor{red!30}50\%  & 90.52 & 79.14 & 88.96 & 88.75 \\
& \cellcolor{orange!25}25\% & \cellcolor{red!45}75\%  & 91.23 & 81.77 & 89.20 & 89.40 \\

\hline
\textit{Glance (Full)} & \cellcolor{gray!10}0\% & \cellcolor{red!60}100\% & \textbf{91.96} & \textbf{84.94} & \textbf{89.40} & \textbf{89.85} \\

\hline\hline
\end{tabular}
}
\end{table}

%% file: tables/ablation_gaussian_kernel.tex
\begin{table}[!t]
\centering
\caption{Ablation study of different kernel representation.}
\label{tab:gaussian_kernel}

\resizebox{0.8\textwidth}{!}{
\begin{tabular}{>{\centering}p{3cm}c|cc|cc}
\hline\hline
\multirow{2}{*}{Distribution} & \multirow{2}{*}{Kernel} & \multicolumn{2}{c|}{UCF-Crime} & \multicolumn{2}{c}{XD-Violence} \\
& & \textbf{AUC(\%)} & \textbf{AUC$_A$(\%)} & \textbf{AP(\%)}$\uparrow$ & \textbf{AP$_A$(\%)}$\uparrow$ \\
\hline
\textit{Normal} (Default) & \raisebox{-.5\height}{\includegraphics[width=2cm]{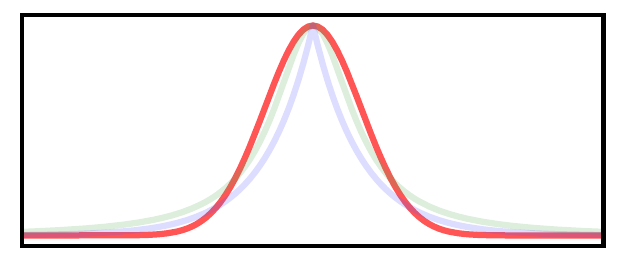}} & \textbf{91.96} & \textbf{84.94} & \textbf{89.40} & \textbf{89.85} \\
\hline
\textit{Cauchy} & \raisebox{-.5\height}{\includegraphics[width=2cm]{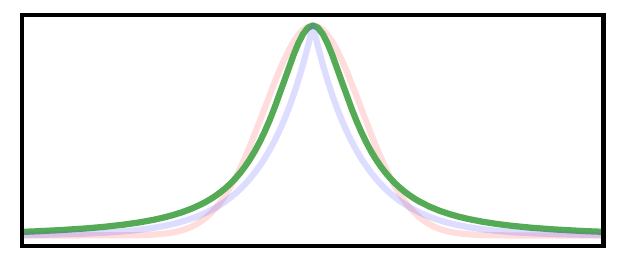}} & 91.38 & 82.42 & 88.24 & 89.06 \\
\hline
\textit{Laplace} & \raisebox{-.5\height}{\includegraphics[width=2cm]{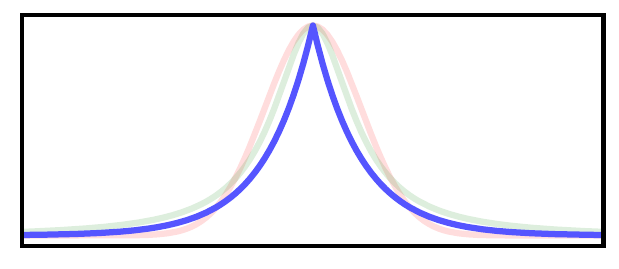}} & 90.86 & 81.75 & 88.98 & 89.53 \\
\hline\hline
\end{tabular}
}

\end{table}

%% file: figs/quality_result.tex
\begin{figure*}[t!]
\centering
\includegraphics[scale=0.2]{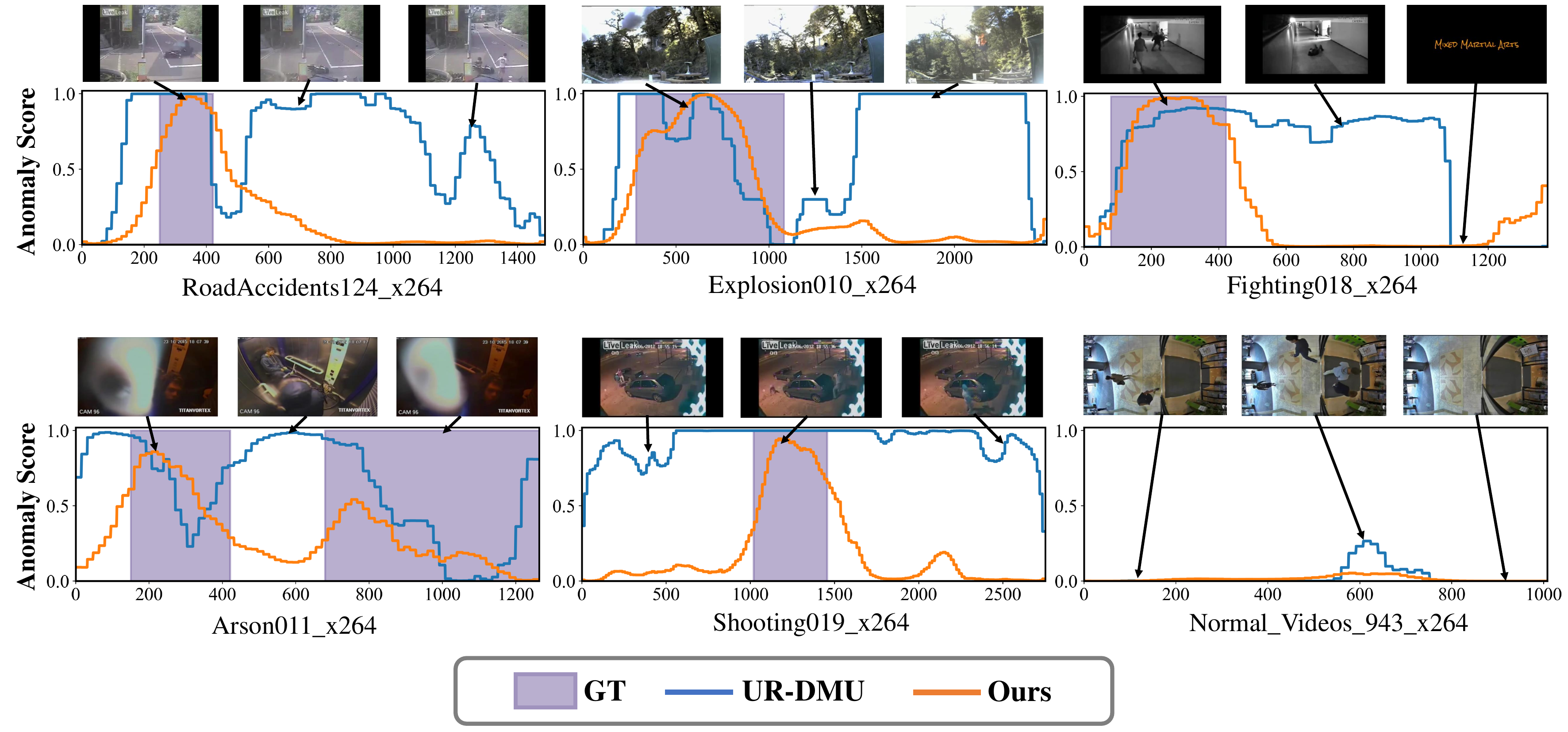}
\vspace{-4mm}
\caption{Qualitative comparison of the baseline method (UR-DMN) and our method on UCF-Crime dataset.
}
\label{fig:quality}
\end{figure*}

%% file: src/5-conclusion.tex
\section{Conclusion}
In this paper, we introduce a novel labeling paradigm, namely glance annotation, for video anomaly detection.
%
To evaluate the effectiveness of glance supervision, we manually annotate glances in the two largest datasets: UCF-Crime and XD-Violence.
%
We present \textbf{GlanceVAD}, which adopts our proposed Temporal Gaussian Splatting to exploit smooth and dense guidance from glance annotations.
Implemented across various MIL-based networks, our approach significantly outperforms baselines and previous state-of-the-art methods and exhibits an efficient trade-off between annotation cost and anomaly detection performance.
%

\section{Limitation and Future Work}
\vspace{-2mm}
Our current work focuses solely on annotating videos in the temporal dimension, leaving out the exploration of valuable spatial information. Investigating the application of glance annotations to the spatial dimension is an important avenue for future research.
Furthermore, the current weakly supervised video anomaly detection setting is restricted to the closed-set scenario in this study. We aspire to extend our glance-based annotation paradigm to more generic open-set scenarios, enabling its utilization in interactive video anomaly detection.

%% file: src/6-acknowledgement.tex
\noindent\textbf{Acknowledgement}
This work is supported by the National Natural Science Foundation of China under grant U22B2053.

%% file: supply/appendix.tex
\setcounter{table}{0}  
\setcounter{figure}{0}  
\renewcommand{\thetable}{\Alph{table}}
\renewcommand{\thefigure}{\Alph{figure}}

\appendix

\input{supply/interface}
\section{Glance Annotation Process}
\subsection{Annotation tool}
We develop a interface for the purpose of glance annotation in videos, as illustrated in \cref{interface}.
The interface facilitates the reading of video lists, quickly adjusting video progress, and automatically recording timestamps for annotating single-frame.
Additionally, the interface allows for the preview of annotated frames, by clicking the annotated frame ID, the video progress automatically synchronizes with the corresponding temporal position.
These features streamline the entire annotation process, enhancing both convenience and efficiency.
In case annotators detect errors or necessitate adjustments, they can delete incorrect annotations and proceed with re-annotation.

\subsection{Quality control}
We first divide the entire dataset into different portions and distributed them to different annotators for labeling.
After completing the first round of annotations,
we conduct a secondary review of the video annotations to remove wrong/redundant annotations.
Additionally, we add clicks which are ignored by annotators to minimize omissions of possible anomalies.
This process ensure the \textbf{Reliability} and \textbf{Comprehensiveness} of the glance annotations.

\subsection{Examples of glance annotation}
We provide several examples of annotated videos in \cref{glance_examples} for better understanding of the annotation process.
\input{supply/glance_examples}

\section{A Revisit of Baseline Methods}
In the following section, we revisit our baseline methods, namely, MIL~\cite{ucf}, RTFM~\cite{rtfm}, and UR-DMU~\cite{URDMU}, and present a comparison in \cref{baseline_methods}.
Note that our method is adaptable to any MIL-based networks, we choose the mentioned methods as our baselines due to their straightforward implementation and outstanding performance.
\subsection{MIL}
We follow~\cite{ucf} to construct our MIL baseline, which consists of a visual encoder and a score head, both implemented with one-dimensional convolutions.
The loss function of MIL baseline includes only video-level BCE loss.
Therefore, the base loss function of MIL baseline is:
\begin{equation}
{\mathcal L}_{base} = {\mathcal L}_{mil}
\end{equation}

\subsection{RTFM}
RTFM~\cite{rtfm} employs a Multi-scale Temporal Network (MTN) as the visual encoder and constructs the score head using multiple layers of Linear operations.
In addition to the basic video-level BCE loss, RTFM also introduces the feature magnitude loss ${\mathcal L}_{mag}$, which is based on the difference in magnitude between features of normal and abnormal video snippets.
This is aimed at maximizing the separability between normal and abnormal videos.
The base loss function of RTFM baseline is:
\begin{equation}
{\mathcal L}_{base} = {\mathcal L}_{mil} + {\mathcal L}_{mag}
\end{equation}

\subsection{UR-DMU}
UR-DMU~\cite{URDMU} use a Global and Local Multi-Head Self Attention (GL-MHSA) module as the visual encoder, which aims to capture the long-range and short-range temporal relationships between video snippets.
Additionally, UR-DMU introduce two memory banks to store and separate abnormal and normal prototypes and maximize the margins between the two representations.
To learn discriminative representations, UR-DMU employs triplet loss to increase the feature distance after interacting with different memories. Simultaneously, it uses KL loss to constrain normal memory to follow a Gaussian distribution, covering the variance introduced by noise.
Hence, the base loss function of UR-DMU baseline is:
\begin{equation}
{\mathcal L}_{base} = {\mathcal L}_{mil} + {\mathcal L}_{mag} + {\mathcal L}_{triplet} + {\mathcal L}_{kl}
\end{equation}

\input{supply/baseline_models}

\section{More Qualitative Results}
In \cref{fig:quality_supply}, we show additional qualitative results on the UCF-Crime dataset and XD-Violence dataset, providing further evidence of the enhanced anomaly detection capabilities of our method.
\input{supply/quality_result_supply}

%% file: supply/interface.tex
\begin{figure}[h]
\centering
\includegraphics[scale=0.4]{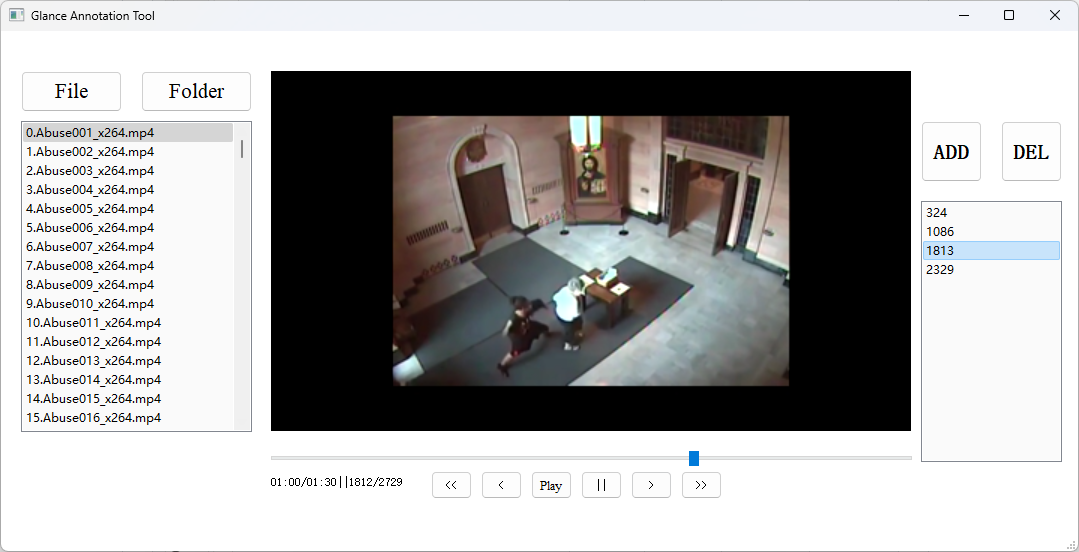}
\caption{\textbf{Screenshot of the glance annotation interface}.
}
\label{interface}
\end{figure}

%% file: supply/glance_examples.tex
\begin{figure}[t]
\centering
\includegraphics[scale=0.32]{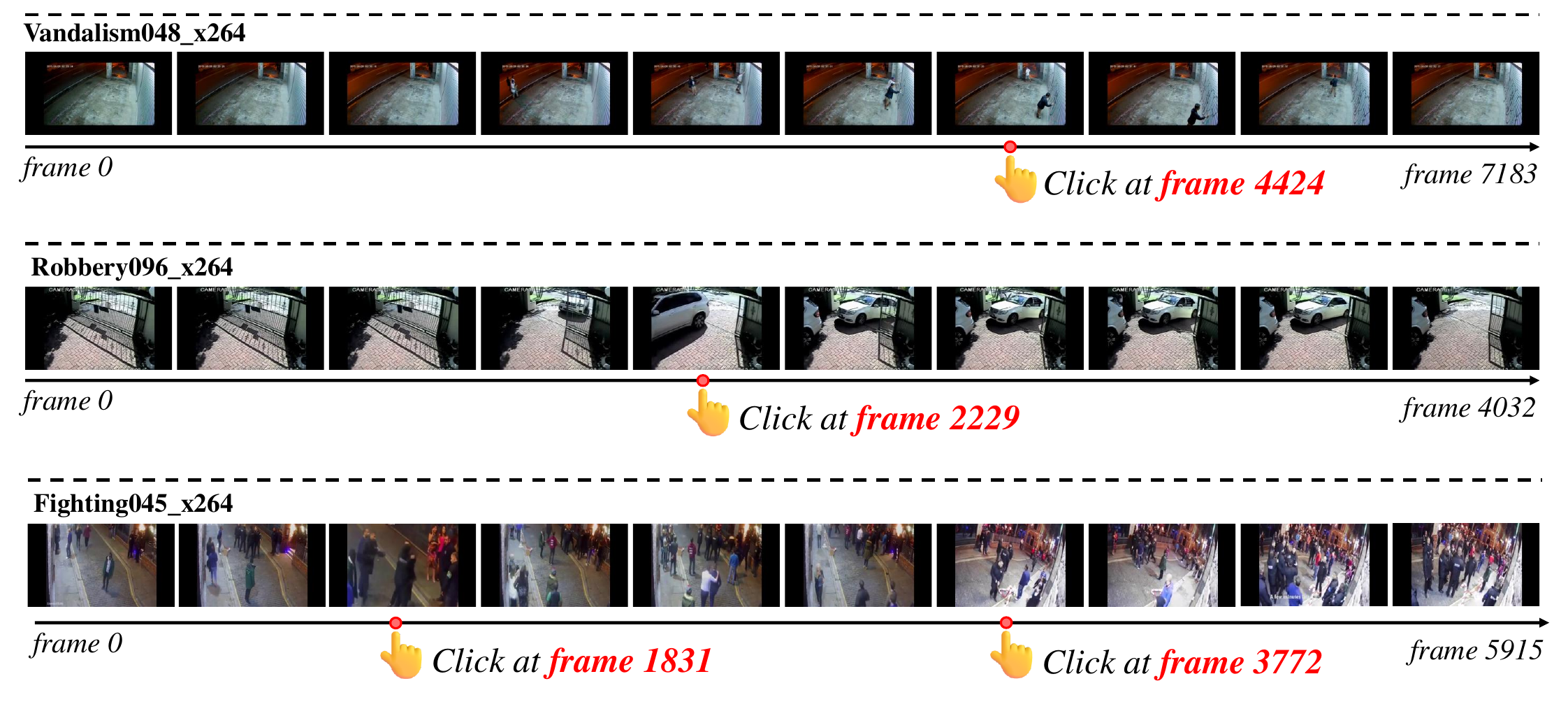}
\caption{\textbf{Examples of Glance Annotation}.
}
\label{glance_examples}
\end{figure}

%% file: supply/baseline_models.tex
\begin{figure}[t]
\centering
\includegraphics[scale=0.45]{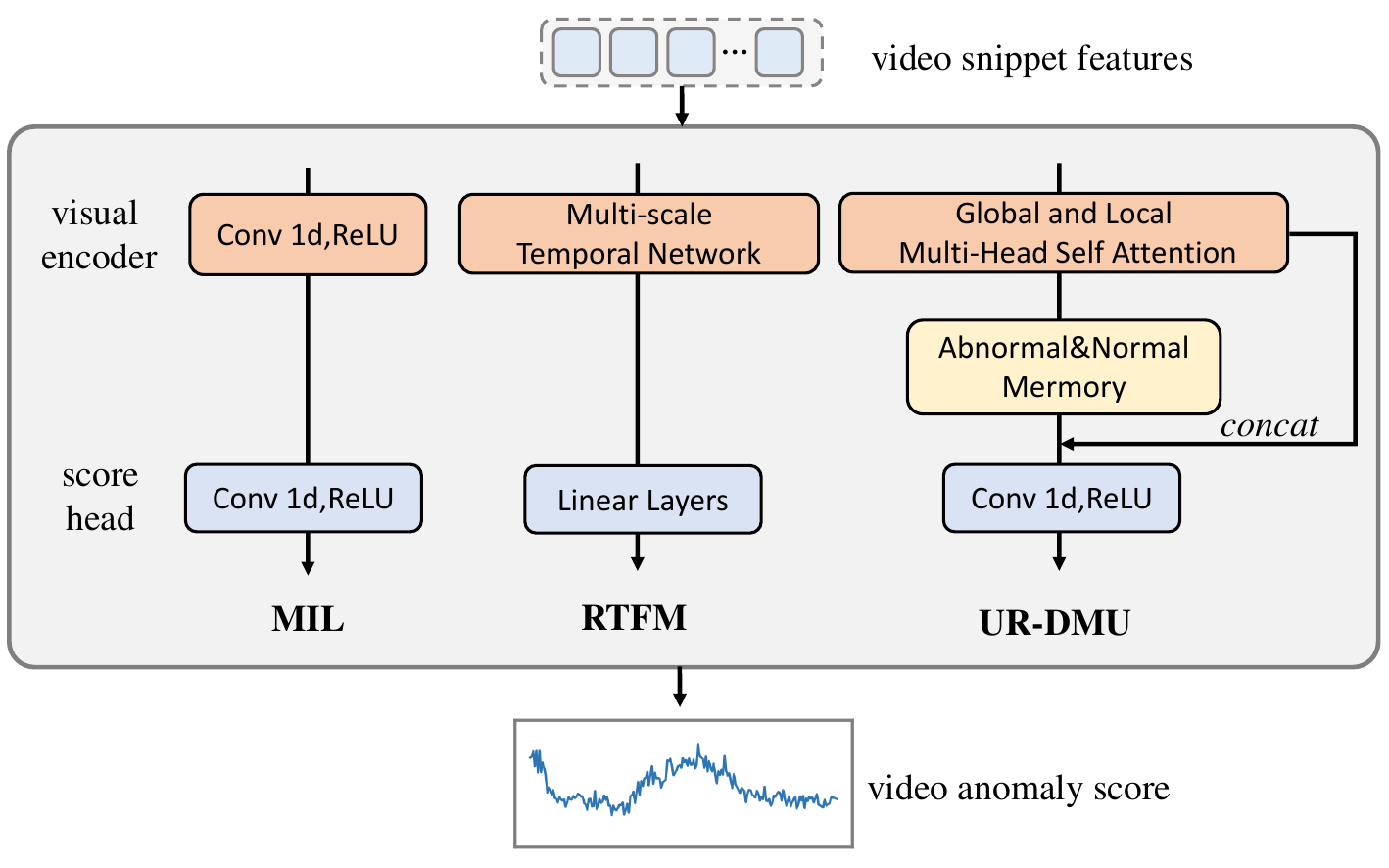}
\caption{\textbf{Comparison of baseline methods}.
}
\label{baseline_methods}
\end{figure}

%% file: supply/quality_result_supply.tex
\begin{figure}[h]
\centering
\includegraphics[scale=0.2]{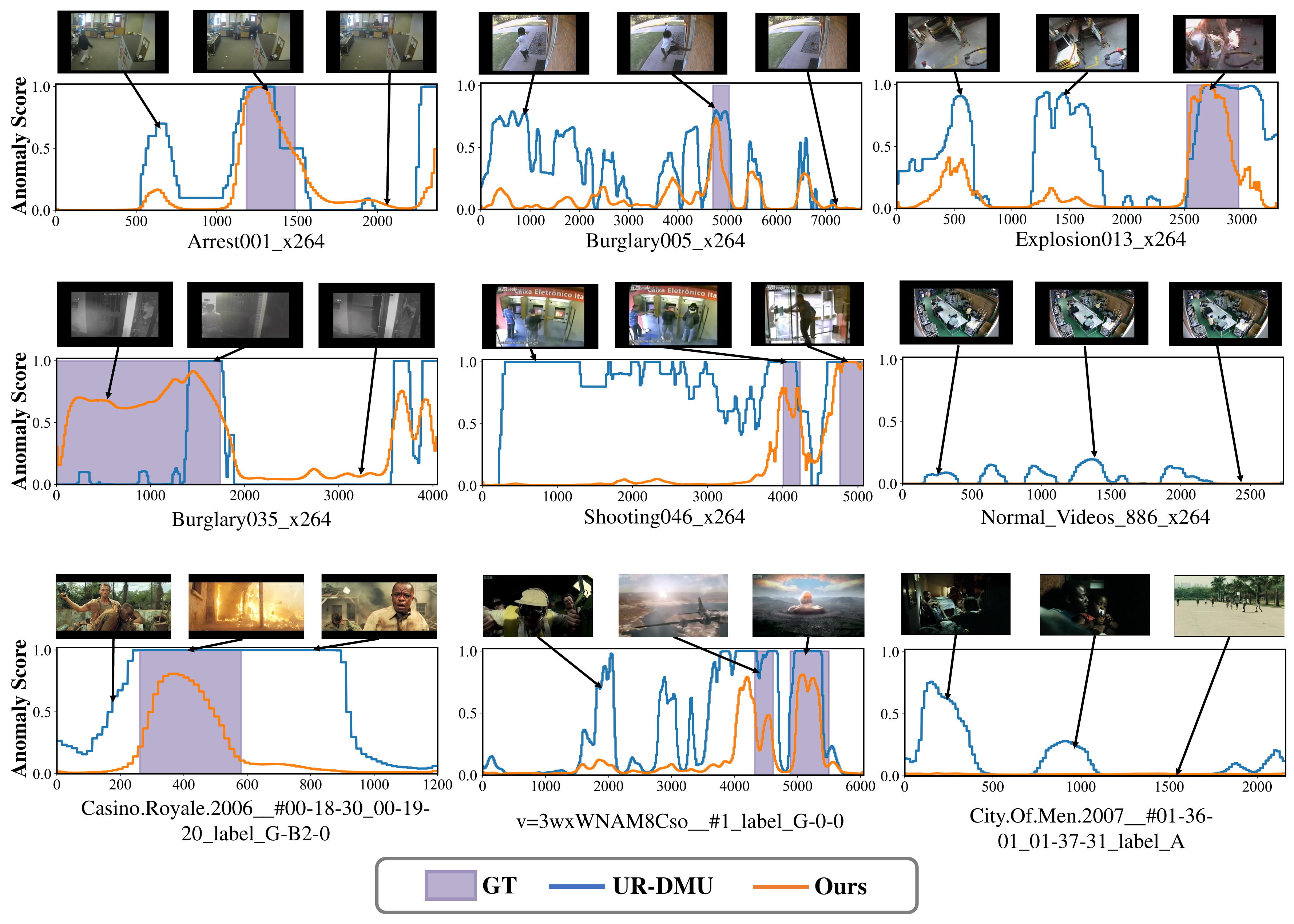}
\vspace{-4mm}
\caption{Qualitative comparison of the baseline method (UR-DMU) and our method on UCF-Crime dataset and XD-Violence dataset.
}
\label{fig:quality_supply}
\end{figure}

%% file: main.bbl
\begin{thebibliography}{10}
\providecommand{\url}[1]{\texttt{#1}}
\providecommand{\urlprefix}{URL }
\providecommand{\doi}[1]{https://doi.org/#1}

\bibitem{adam2008robust}
Adam, A., Rivlin, E., Shimshoni, I., Reinitz, D.: Robust real-time unusual event detection using multiple fixed-location monitors. IEEE transactions on pattern analysis and machine intelligence  \textbf{30}(3),  555--560 (2008)

\bibitem{bearman2016s}
Bearman, A., Russakovsky, O., Ferrari, V., Fei-Fei, L.: What’s the point: Semantic segmentation with point supervision. In: European conference on computer vision. pp. 549--565. Springer (2016)

\bibitem{generic_ad}
Cao, Y., Xu, X., Sun, C., Huang, X., Shen, W.: Towards generic anomaly detection and understanding: Large-scale visual-linguistic model (gpt-4v) takes the lead. arXiv preprint arXiv:2311.02782  (2023)

\bibitem{cao2024survey}
Cao, Y., Xu, X., Zhang, J., Cheng, Y., Huang, X., Pang, G., Shen, W.: A survey on visual anomaly detection: Challenge, approach, and prospect. arXiv preprint arXiv:2401.16402  (2024)

\bibitem{i3d}
Carreira, J., Zisserman, A.: Quo vadis, action recognition? a new model and the kinetics dataset. In: CVPR. pp. 6299--6308 (2017)

\bibitem{cui2022video}
Cui, R., Qian, T., Peng, P., Daskalaki, E., Chen, J., Guo, X., Sun, H., Jiang, Y.G.: Video moment retrieval from text queries via single frame annotation. In: Proceedings of the 45th International ACM SIGIR Conference on Research and Development in Information Retrieval. pp. 1033--1043 (2022)

\bibitem{mist}
Feng, J.C., Hong, F.T., Zheng, W.S.: Mist: Multiple instance self-training framework for video anomaly detection. In: Proceedings of the IEEE/CVF conference on computer vision and pattern recognition. pp. 14009--14018 (2021)

\bibitem{gong2019memorizing}
Gong, D., Liu, L., Le, V., Saha, B., Mansour, M.R., Venkatesh, S., Hengel, A.v.d.: Memorizing normality to detect anomaly: Memory-augmented deep autoencoder for unsupervised anomaly detection. In: Proceedings of the IEEE/CVF International Conference on Computer Vision. pp. 1705--1714 (2019)

\bibitem{ConvAE}
Hasan, M., Choi, J., Neumann, J., Roy-Chowdhury, A.K., Davis, L.S.: Learning temporal regularity in video sequences. In: Proceedings of the IEEE conference on computer vision and pattern recognition. pp. 733--742 (2016)

\bibitem{huang2018makes}
Huang, D.A., Ramanathan, V., Mahajan, D., Torresani, L., Paluri, M., Fei-Fei, L., Niebles, J.C.: What makes a video a video: Analyzing temporal information in video understanding models and datasets. In: Proceedings of the IEEE Conference on Computer Vision and Pattern Recognition. pp. 7366--7375 (2018)

\bibitem{kerbl20233d}
Kerbl, B., Kopanas, G., Leimk{\"u}hler, T., Drettakis, G.: 3d gaussian splatting for real-time radiance field rendering. ACM Transactions on Graphics  \textbf{42}(4) (2023)

\bibitem{keselman2023flexible}
Keselman, L., Hebert, M.: Flexible techniques for differentiable rendering with 3d gaussians. arXiv preprint arXiv:2308.14737  (2023)

\bibitem{kim2009observe}
Kim, J., Grauman, K.: Observe locally, infer globally: a space-time mrf for detecting abnormal activities with incremental updates. In: 2009 IEEE conference on computer vision and pattern recognition. pp. 2921--2928. IEEE (2009)

\bibitem{landi2019anomaly}
Landi, F., Snoek, C.G., Cucchiara, R.: Anomaly locality in video surveillance. arXiv preprint arXiv:1901.10364  (2019)

\bibitem{lacp}
Lee, P., Byun, H.: Learning action completeness from points for weakly-supervised temporal action localization. In: ICCV. pp. 13648--13657 (2021)

\bibitem{li2023d3g}
Li, H., Shu, X., He, S., Qiao, R., Wen, W., Guo, T., Gan, B., Sun, X.: D3g: Exploring gaussian prior for temporal sentence grounding with glance annotation. arXiv preprint arXiv:2308.04197  (2023)

\bibitem{MSL}
Li, S., Liu, F., Jiao, L.: Self-training multi-sequence learning with transformer for weakly supervised video anomaly detection. In: Proceedings of the AAAI Conference on Artificial Intelligence. vol.~36, pp. 1395--1403 (2022)

\bibitem{li2013anomaly}
Li, W., Mahadevan, V., Vasconcelos, N.: Anomaly detection and localization in crowded scenes. IEEE transactions on pattern analysis and machine intelligence  \textbf{36}(1),  18--32 (2013)

\bibitem{li2021temporal}
Li, Z., Abu~Farha, Y., Gall, J.: Temporal action segmentation from timestamp supervision. In: Proceedings of the IEEE/CVF Conference on Computer Vision and Pattern Recognition. pp. 8365--8374 (2021)

\bibitem{liu2019exploring}
Liu, K., Ma, H.: Exploring background-bias for anomaly detection in surveillance videos. In: Proceedings of the 27th ACM International Conference on Multimedia. pp. 1490--1499 (2019)

\bibitem{framepred}
Liu, W., Luo, W., Lian, D., Gao, S.: Future frame prediction for anomaly detection--a new baseline. In: Proceedings of the IEEE conference on computer vision and pattern recognition. pp. 6536--6545 (2018)

\bibitem{Luetal}
Lu, C., Shi, J., Jia, J.: Abnormal event detection at 150 fps in matlab. In: Proceedings of the IEEE international conference on computer vision. pp. 2720--2727 (2013)

\bibitem{luo2023frequency}
Luo, R., Wei, R., Gao, C., Sang, N.: Frequency information matters for image matting. In: Asian Conference on Pattern Recognition. pp. 81--94. Springer (2023)

\bibitem{rnn}
Luo, W., Liu, W., Gao, S.: A revisit of sparse coding based anomaly detection in stacked rnn framework. In: Proceedings of the IEEE international conference on computer vision. pp. 341--349 (2017)

\bibitem{UMIL}
Lv, H., Yue, Z., Sun, Q., Luo, B., Cui, Z., Zhang, H.: Unbiased multiple instance learning for weakly supervised video anomaly detection. In: Proceedings of the IEEE/CVF Conference on Computer Vision and Pattern Recognition. pp. 8022--8031 (2023)

\bibitem{sf}
Ma, F., Zhu, L., Yang, Y., Zha, S., Kundu, G., Feiszli, M., Shou, Z.: Sf-net: Single-frame supervision for temporal action localization. In: ECCV. pp. 420--437. Springer (2020)

\bibitem{mehran2009abnormal}
Mehran, R., Oyama, A., Shah, M.: Abnormal crowd behavior detection using social force model. In: 2009 IEEE conference on computer vision and pattern recognition. pp. 935--942. IEEE (2009)

\bibitem{mettes2016spot}
Mettes, P., Van~Gemert, J.C., Snoek, C.G.: Spot on: Action localization from pointly-supervised proposals. In: Computer Vision--ECCV 2016: 14th European Conference, Amsterdam, The Netherlands, October 11-14, 2016, Proceedings, Part V 14. pp. 437--453. Springer (2016)

\bibitem{unet}
Ronneberger, O., Fischer, P., Brox, T.: U-net: Convolutional networks for biomedical image segmentation. In: Medical Image Computing and Computer-Assisted Intervention--MICCAI 2015: 18th International Conference, Munich, Germany, October 5-9, 2015, Proceedings, Part III 18. pp. 234--241. Springer (2015)

\bibitem{OCSVM}
Sch{\"o}lkopf, B., Williamson, R.C., Smola, A., Shawe-Taylor, J., Platt, J.: Support vector method for novelty detection. Advances in neural information processing systems  \textbf{12} (1999)

\bibitem{ucf}
Sultani, W., Chen, C., Shah, M.: Real-world anomaly detection in surveillance videos. In: Proceedings of the IEEE conference on computer vision and pattern recognition. pp. 6479--6488 (2018)

\bibitem{stcgraph}
Sun, C., Jia, Y., Hu, Y., Wu, Y.: Scene-aware context reasoning for unsupervised abnormal event detection in videos. In: Proceedings of the 28th ACM International Conference on Multimedia. pp. 184--192 (2020)

\bibitem{rtfm}
Tian, Y., Pang, G., Chen, Y., Singh, R., Verjans, J.W., Carneiro, G.: Weakly-supervised video anomaly detection with robust temporal feature magnitude learning. In: Proceedings of the IEEE/CVF international conference on computer vision. pp. 4975--4986 (2021)

\bibitem{wang2022voge}
Wang, A., Wang, P., Sun, J., Kortylewski, A., Yuille, A.: Voge: a differentiable volume renderer using gaussian ellipsoids for analysis-by-synthesis. In: The Eleventh International Conference on Learning Representations (2022)

\bibitem{GODs}
Wang, J., Cherian, A.: Gods: Generalized one-class discriminative subspaces for anomaly detection. In: Proceedings of the IEEE/CVF International Conference on Computer Vision. pp. 8201--8211 (2019)

\bibitem{wang2023molo}
Wang, X., Zhang, S., Qing, Z., Gao, C., Zhang, Y., Zhao, D., Sang, N.: Molo: Motion-augmented long-short contrastive learning for few-shot action recognition. In: Proceedings of the IEEE/CVF Conference on Computer Vision and Pattern Recognition. pp. 18011--18021 (2023)

\bibitem{S3R}
Wu, J.C., Hsieh, H.Y., Chen, D.J., Fuh, C.S., Liu, T.L.: Self-supervised sparse representation for video anomaly detection. In: European Conference on Computer Vision. pp. 729--745. Springer (2022)

\bibitem{xdviolence}
Wu, P., Liu, J., Shi, Y., Sun, Y., Shao, F., Wu, Z., Yang, Z.: Not only look, but also listen: Learning multimodal violence detection under weak supervision. In: Computer Vision--ECCV 2020: 16th European Conference, Glasgow, UK, August 23--28, 2020, Proceedings, Part XXX 16. pp. 322--339. Springer (2020)

\bibitem{vadclip}
Wu, P., Zhou, X., Pang, G., Zhou, L., Yan, Q., Wang, P., Zhang, Y.: Vadclip: Adapting vision-language models for weakly supervised video anomaly detection. arXiv preprint arXiv:2308.11681  (2023)

\bibitem{xu2022reliable}
Xu, X., Wang, J., Li, X., Lu, Y.: Reliable propagation-correction modulation for video object segmentation. In: Proceedings of the AAAI Conference on Artificial Intelligence. vol.~36, pp. 2946--2954 (2022)

\bibitem{yang2023video}
Yang, Z., Liu, J., Wu, Z., Wu, P., Liu, X.: Video event restoration based on keyframes for video anomaly detection. In: Proceedings of the IEEE/CVF Conference on Computer Vision and Pattern Recognition. pp. 14592--14601 (2023)

\bibitem{MACILSD}
Yu, J., Liu, J., Cheng, Y., Feng, R., Zhang, Y.: Modality-aware contrastive instance learning with self-distillation for weakly-supervised audio-visual violence detection. In: Proceedings of the 30th ACM International Conference on Multimedia. pp. 6278--6287 (2022)

\bibitem{hrpro}
Zhang, H., Wang, X., Xu, X., Qing, Z., Gao, C., Sang, N.: Hr-pro: Point-supervised temporal action localization via hierarchical reliability propagation. arXiv preprint arXiv:2308.12608  (2023)

\bibitem{zhao2011online}
Zhao, B., Fei-Fei, L., Xing, E.P.: Online detection of unusual events in videos via dynamic sparse coding. In: CVPR 2011. pp. 3313--3320. IEEE (2011)

\bibitem{GCN}
Zhong, J.X., Li, N., Kong, W., Liu, S., Li, T.H., Li, G.: Graph convolutional label noise cleaner: Train a plug-and-play action classifier for anomaly detection. In: Proceedings of the IEEE/CVF conference on computer vision and pattern recognition. pp. 1237--1246 (2019)

\bibitem{URDMU}
Zhou, H., Yu, J., Yang, W.: Dual memory units with uncertainty regulation for weakly supervised video anomaly detection. arXiv preprint arXiv:2302.05160  (2023)

\end{thebibliography}
